\documentclass[pmlr,twocolumn,10pt]{jmlr} 

\usepackage{booktabs}
\usepackage{siunitx}

\usepackage[switch]{lineno}

\theorembodyfont{\upshape}
\theoremheaderfont{\scshape}
\theorempostheader{:}
\theoremsep{\newline}

\jmlrvolume{333}
\jmlryear{2026}
\jmlrworkshop{Conference on Health, Inference, and Learning (CHIL) 2026} 

\title[Extubation failure prediction]{Enhancing Extubation Failure Prediction with LLM-Derived Features from Respiratory Therapy Clinical Notes}

\author{%
    \Name{Izzy Chaiken}
    \Email{chaiken@uw.edu}\\
    \addr Information School, University of Washington, Seattle, WA, USA
    \AND
    \Name{Aditya Khowal}
    \Email{akhowal@uw.edu}\\
    \addr Computer Science \& Engineering, University of Washington, Seattle, WA, USA
    \AND
    \Name{Neha A. Sathe}
    \Email{nas212@uw.edu}\\
    \addr Department of Medicine, University of Washington, Seattle, WA, USA
    \AND
    \Name{Mark M. Wurfel}
    \Email{mwurfel@uw.edu}\\
    \addr Department of Medicine, University of Washington, Seattle, WA, USA
    \AND
    \Name{Lucy Lu Wang}
    \Email{lucylw@uw.edu}\\
    \addr Information School, University of Washington, Seattle, WA, USA
}

\usepackage{graphicx} 

\usepackage[load-configurations=version-1]{siunitx} 
\usepackage{multirow}
\usepackage{makecell}

\usepackage{amssymb}
\usepackage{tikz}
\usepackage{booktabs}
\usepackage{longtable}
\usepackage{subcaption}

\usepackage{xspace}
\usepackage{xcolor}
\usepackage{enumitem}
\usepackage{hyperref}
\usepackage{mathtools}
\usepackage{caption}
\usepackage{arydshln}
\usepackage{bold-extra}
\usepackage{soul}
\usepackage[most]{tcolorbox}

\newcommand\uwname{University of Washington Medicine\xspace}

\newcolumntype{L}[1]{>{\raggedright\let\newline\\\arraybackslash\hspace{0pt}}p{#1}}
\newcolumntype{M}[1]{>{\raggedright\let\newline\\\arraybackslash\hspace{0pt}}m{#1}}
\newcolumntype{C}[1]{>{\centering\arraybackslash\hspace{0pt}}p{#1}}

\definecolor{darkgreen}{rgb}{0.0, 0.4, 0.13}

\begin{document}

\maketitle

\begin{abstract}
Invasive mechanical ventilation is a lifesaving therapy, but timely, safe discontinuation is essential to preventing extubation failure (EF) and related risks to health. 
We present a novel approach to EF prediction that leverages features
classified in free-text respiratory therapy notes using a large language model and logistic regression pipeline.
Applied to a patient cohort from \uwname, our method identifies clinically meaningful EF-related features that improve EF prediction performance when included alongside structured patient data.
We further highlight how differences in target populations in prior EF prediction studies, such as heterogenous inclusion criteria and EF definition, can lead to systematic differences in model performance and hinder generalizability between studies.
\end{abstract}
\begin{keywords}
    extubation failure, clinical outcome prediction, large language models, EHR
\end{keywords}

\paragraph*{Data and Code Availability}
We constructed a novel electronic health records (EHR) dataset from patients at \uwname. Our dataset includes adult patients who received invasive mechanical ventilation across 10,810 visits to any of the three hospitals of \uwname between April 2021 and September 2023. We utilized both these patients' structured data and unstructured respiratory therapy notes. Personally identifiable information (PII) from our dataset was stored on a HIPAA-compliant server and only accessed by individuals with relevant certification. This dataset has not be made publicly available as it includes PII. Code for reproducing experiments is available at \href{https://github.com/larchlab/extubation-failure-camera-ready}{https://github.com/larchlab/extubation-failure-camera-ready}.

\begin{figure}[t]
    \centering
    \includegraphics[width=0.88\linewidth]{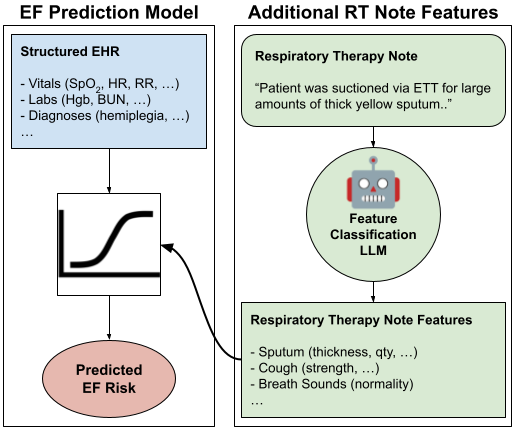}%
    \caption{Our EF outcome prediction pipeline. Models developed in prior work include only features derived from tabular EHR (left). Our models include features classified from clinical notes using LLMs to produce more complete risk assessments (right).}%
    \label{fig:pipeline}%
\end{figure}

\paragraph*{Institutional Review Board (IRB)}
This work is approved by the IRB of the Human Subjects Division of University of Washington under protocol number STUDY00018582.


\section{Introduction}

Invasive mechanical ventilation (IMV) is a lifesaving therapy for patients with respiratory failure
\citep{wunsch2010epidemiology, mehta2015epidemiological}. 
The decision to extubate (discontinue IMV) is complex and balances the costs of ongoing IMV against the risk of \emph{extubation failure} (EF), when a patient dies or requires re-intubation following extubation \citep{thille2013decision}. Prolonged IMV is associated with complications such as airway injury, pneumonia, and
long-term functional deficits
\citep{herridge2011functional, herridge2003one, jubran2019long, klompas2015preventability}, while EF itself is associated with prolonged IMV, longer ICU stays, and excess mortality \citep{thille2011outcomes, epstein1997effect, beduneau2017epidemiology, frutos2011outcome}. 
Physicians vary in deciding when to extubate and how to use therapies to reduce the risk of extubation failure \citep{ely1996effect, betbese1998prospective, wennberg2011time}.

To support these decisions and help reduce EF, clinicians may use clinical decision support systems (CDSS) to augment and standardize human judgment. These CDSS may be developed through prospective clinical trials, which directly test whether specific risk factors are related to extubation outcomes \citep{burns2025liberation}.
For example, \uwname employs a custom CDSS to assess EF risk, and patients found to be at high risk for EF receive additional evaluation and monitoring; in the cohort we examine, roughly half of patients assessed with this tool were classified as high risk and half as low risk,
yet these groups had similar EF rates, 
highlighting inefficiencies in current treatment allocation and opportunities to improve EF prediction accuracy.

\begin{figure*}[t!]
    \centering
    \includegraphics[width=0.9\linewidth]{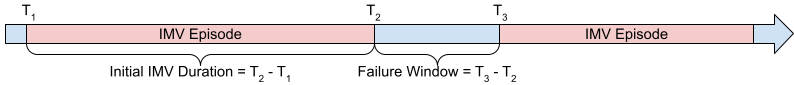}%
    \caption{Example clinical time course for patient experiencing EF. Minimum initial IMV duration and maximum time to failure (i.e., the \emph{failure window}) are used to define cohorts for modeling and whether a patient experiences EF. We assess the effects of varying these two criteria on downstream performance. Because of the frequency of variable collection, structured model features from the 4 hours prior to the end of the initial IMV episode and note features from 12 hours prior to the end of the initial IMV episode are included in the EF prediction models.}%
    \label{fig:illustrative_example}%
\end{figure*}

Prior research applies machine learning techniques to predict EF, but these models are restricted to using inputs available in tabular electronic health record (EHR) data (e.g. vitals, labs, or ventilator information) \citep{igarashi2022machine}.
These models do not exploit the rich information in clinical notes collected during IMV, which encode patients' airway and respiratory status.
Recent research has demonstrated the feasibility of applying large language models (LLMs) to generate outcome predictions either directly from unstructured notes \citep{van2021clinical, naik2022literature}, or to extract entities from notes at scale for incorporation into predictive models \citep{robitschek2025large, mugisha2022comparison}.
Building on clinical information extraction and EF prediction literature, we define and design a novel LLM-based pipeline to classify features in respiratory therapy (RT) notes.\footnote{These notes are distinct from those found in publicly available EHR datasets such as MIMIC-IV \citep{johnson2023mimic}, as they include descriptions of a patient's state collected during IMV.} We develop EF prediction models including these features, and demonstrate that they improve prediction performance, revealing new risk factors that are important to document and model.


Prior EF research also uses inconsistent definitions for inclusion criteria. \autoref{fig:illustrative_example} illustrates an example clinical time course for a patient who experiences EF: an initial IMV episode begins at time $T_1$ and ends with extubation at $T_2$, before the patient experiences EF at $T_3$ when IMV is reapplied. Patient cohorts for EF risk investigations use a \emph{minimum IMV duration} ($T_2 - T_1$) to define valid initial IMV episodes for inclusion. Failure events are then labeled according to a \emph{maximum failure window}---EF occurs if failure is within this window but not after
(i.e., if $T_3 - T_2$ is greater than this window, the patient will be labeled negative for EF, despite experiencing reintubation).
Patients with different initial IMV durations exhibit variable EF risk profiles \citep{thille2019effect}, 
impeding comparability of models trained using different cohort definitions and hindering consistent interpretations of EF risk factors \citep{nava2005noninvasive}. \autoref{litreview} shows this inconsistency in definitions adopted by prior work.
For example, most modeling studies define EF as occurring within 2-3 days of extubation \citep{torrini2021prediction}, whereas prospective clinical studies instead use a 7-day window \citep{beduneau2017epidemiology}.
To focus our modeling on high risk patients and ensure all clinically relevant EF is labeled as such, we estimate EF risk among patients with IMV duration of at least 24 hours and define failure as occurring within 7 days of extubation. We further train models with varying EF definitions and inclusion criteria and show their impacts on model performance and interpretation. 

Our contributions are as follows:


\begin{itemize}[itemsep=-1pt, topsep=0pt, leftmargin=12pt]
    \item We introduce a schema of 15 features described in unstructured respiratory therapy (RT) notes that are relevant to EF, e.g., cough strength and sputum quantity. We develop LLM-based classifiers that reliably identify presence of these features; we validate against a set of 200 RT notes manually annotated for feature values (\S\ref{sec:rtnotes}).
    \item We construct a cohort of 10k+ patients who underwent IMV at \uwname (\S\ref{sec:data}), and incorporate their RT note features into EF risk models (\S\ref{sec:ef_methods}). We find that RT note features improve AUROC by 1.9 points for our best performing models, with a larger difference when models are trained only over patients with documented RT notes (5.1 points) (\S\ref{sec:results}).
    \item We investigate how different 
    inclusion criteria and EF definitions impact model performance (\S\ref{sec:results}). Varying minimum IMV duration strongly impacts performance metrics: AUROC generally decreases and F1 increases as minimum IMV duration is increased.
    Since performance of EF predictors varies with characteristics of the patient population, models trained on cohorts with different inclusion criteria are not directly comparable.
\end{itemize}


\vspace{-3mm}
\section{Related Work}
\label{sec:related_work}
\vspace{-1mm}

\begin{table*}[h!]
    \small
    \centering
    \begin{tabular}{llcccc}
    \toprule
        Source & Model Type & Cohort Size & EF Rate & Failure Window (days) & Min Intub Len (hrs) \\
        \midrule
        \citet{seely2014heart} & LR ensemble & 434 & 12\% & 2 & 48 \\
        \citet{hsieh2018artificial} & ANN & 3602 & 5\% & 3 & 0 \\
        \citet{chen2019prediction} & GB & 3636 & 17\% & 2 & 0 \\
        \citet{fabregat2021machine} & SVM & 1108 & 9\% & 7 & 12 \\
        \citet{fleuren2021predictors} & GB & 883 & 19\% & 7 & 24 \\
        \citet{otaguro2021machine} & GB & 117 & 11\% & 3 & 24 \\
        \citet{zhao2021development} & GB & 16189 & 17\% & 2 & 0 \\
        \citet{zeng2022interpretable} & RNN & 8599 & 30\% & 2 & 12 \\
        \midrule
        \textbf{Ours} & LR & 3243 & 20\% & 7 & 24 \\
        \bottomrule
    \end{tabular}
    \vspace{-2mm}
    \caption{Model type, cohort statistics, and inclusion criteria reported in prior literature. EF rate contextualizes the overall outcomes within a patient cohort, providing another indicator of variability within experimental settings}
    \label{litreview}
\end{table*}

\paragraph{Clinical Assessment of EF Risk} Prior studies have developed tools to estimate patient-level EF risk, which guide clinical decisions at the time of planned extubation. If a patient is deemed high-risk for EF, a clinician may defer extubation and/or initiate treatments to reduce risk of failure \citep{thille2019effect, apfelbaum20212022, quintard2019experts, grieco2021non}. 
However, the lack of standardized EF risk assessments leads to variation in assessed risk levels and choice of extubation-time therapies \citep{burns2018international, godard2016practice}. A variety of risk-stratification systems for EF are used in clinical practice
\citep{sarti2021feasibility, joffe2022extubation}, integrating features such as real-time vitals, medical history, and subjective impression of risk \citep{burns2025liberation, joffe2022extubation}. 
Such systems improve patient outcomes during extubation, and similar systems for other aspects of hospital care improve adherence to care guidelines and reduce morbidity \citep{zheng2022economic, moja2014effectiveness}.

\vspace{-1mm}
\paragraph{Machine Learning for EF Prediction} Recent research develops machine learning models to retrospectively predict EF risk in adult ICU patients using structured variables \citep{chen2019prediction, fabregat2021machine, fleuren2021predictors, hsieh2018artificial, otaguro2021machine, seely2014heart, zhao2021development}. Some models also include longitudinal data \citep{seely2014heart, zeng2022interpretable}.
Prior EF prediction models largely utilize gradient boosting (GB) and artificial neural networks (ANN) \citep{igarashi2022machine}, 
and attain AUROC between 0.83-0.85 on variable-sized cohorts \citep{zhao2021development, hsieh2018artificial}. 
However, EF prediction performance may not generalize to external cohorts: 
\citet{zhao2021development} demonstrated their model performance diminishing when tested on a patient cohort from a hospital not included in the training data. Models from prior work are generally unavailable, limiting our ability to assess their performance over our cohort. 

\vspace{-1mm}
\paragraph{EF Definition \& Selection Criteria}
EF is not consistently defined, as the window during which EF may occur varies. 
Therapies such as non-invasive ventilation can delay EF, so shorter EF windows may improperly label patients \citep{thille2016easily}.
No standard minimum initial IMV duration in used in research, so EF risk factors and outcome predictions may be inconsistent \citep{rose2017variation}. IMV duration is itself associated with higher risk of EF, both due to differences in patient population at various durations, and through physiologic effects of prolonged IMV \citep{rothaar2003extubation, torrini2021prediction}. Clinical literature uses IMV duration as a proxy for EF risk, so prior studies differ in patient risk profiles \citep{nava2005noninvasive, thille2013decision}. 
We examine the effects of EF failure windows and cohort selection criteria on EF risk models.


\vspace{-1mm}
\paragraph{Clinical Note Feature Classification}
LLMs have demonstrated high performance in determining information contained in biomedical texts \citep{perera2020named, li2024scoping}, enabling information extraction from clinical notes at scale using few-shot learning techniques \citep{agrawal2022large, goel2023llms}. 
LLMs have been applied to extract features from clinical notes to predict outcomes such as contraceptive switching rationales \citep{miao2025understanding}, bladder cancer survival \citep{sun2024outcome}, concepts related to postpartum hemorrhage \citep{alsentzer2023zero}, and breast cancer phenotypes \citep{zhou2022cancerbert}. We extend this methodology, classifying features related to a patient's IMV status in RT notes.

\vspace{-1mm}
\section{Data}
\label{sec:data}
\vspace{-1mm}

\paragraph{Cohort} As described above, we assemble a cohort of 10,194 patients who received IMV across 10,810 visits to any of the three hospitals of \uwname between April 2021 and September 2023. 
We define \emph{extubation} as the time at which a patient's documented oxygen delivery device changes from IMV to a non-IMV method of delivery, and \emph{extubation failure} as patient death or a return to IMV within 7 days of initial extubation.

Of the initial 10,810 encounters, we drop 843 for having incomplete height/weight information, 4,711 for having no IMV session lasting at least 24 hours, 1,922 for having a `Do Not Intubate/Resuscitate' order within 7 days of extubation, and 91 for being encounters with a non-unique patient. Our primary dataset includes 3,243 IMV sessions from unique patients, of whom 647 (19.95\%) experienced EF within 7 days; 618 of these are due to reintubation, and 29 are due to death. Among patients in the non-excluded cohort, median IMV duration was 60.8 hours, and mean IMV duration was 97.1 hours. The first and third quartiles were 36.6 hours and 112.0 hours. Only 2,339 encounters have corresponding readiness checklists (our clinic's CDSS tool, which queries for the presence of 19 binary factors to assess EF risk) from within 4 hours of extubation. We hold out 646 (20\%) encounters as a test set, based on a random sample stratified by documented patient race/ethnicity and EF outcome. 

To investigate the impact of inclusion criteria, we further include 3,685 encounters from patients not in the aforementioned set whose initial intubation was at least 1 hour, rather than 24. Of these additional patients, 133 (3.61\%) experienced EF. We retain 741 of these patients for our test set, based on the same stratified sampling strategy. See Appendix \ref{tab:demographics} for dataset demographics.

\begin{table}[t!]
    \small
    \centering
    \label{tab:extractmetrics}
    \addtolength{\tabcolsep}{-0.15em}
    {\begin{tabular}{L{21mm}L{21mm}C{29mm}}
    \toprule
    Category & Feature & Macro Rec/Prec/F1 \\
    \midrule
    Sputum & Presence & 0.933/0.943/0.937 \\
     & Thick & 0.988/0.996/0.992 \\
     & Thin & 0.944/0.997/0.969 \\
     & Quantity & 0.881/0.852/0.857 \\
     & Color & 0.876/0.808/0.816 \\
    \midrule
    Cough & Presence & 0.995/0.989/0.991 \\
     & Weak & 0.780/0.825/0.801 \\
     & Strong & 0.935/0.923/0.929 \\
     & Induced & 0.835/0.783/0.806 \\
     & Spontaneous* & 0.888/0.545/0.520 \\
     \midrule
    Suctioning & Presence* & 0.875/0.872/0.873 \\
     & Oral* & 0.713/0.834/0.756 \\
     & Endotracheal* & 0.709/0.798/0.732 \\
     \midrule
    Cuff Leak & Presence & 0.991/0.940/0.963 \\
     \midrule 
    Breath Sounds & Normality & 0.929/0.920/0.924 \\
    \bottomrule
    \end{tabular}}
    \caption{Metrics for features extracted from RT Notes. 
    *variables not included in downstream models due to low F1 or high correlation with other features.}
    \label{tab:extractmetrics}
\end{table}

     
\begin{table*}[t!]
    \small
    \centering
    \begin{tabular}{llll}
    \toprule
    \textbf{Category} & \textbf{Feature} & \textbf{Downstream Feature Mapping} & \textbf{N Mentions} \\
    \midrule
    \textbf{Sputum} & Present & \{Unlabeled: 0, Positive: 1\} & 1734 (53.47\%) \\
     & Consistency & \{Unlabeled: 0, Thin: -1, Thick: 1\} & 748.0 (23.07\%) \\ 
     & Quantity & \{Unlabeled: 0, Low: 1, Medium: 2, High: 3\} & 1478 (45.58\%) \\
     & Color** & \{Unlabeled: 0, Non-Pathological: 0, Pathological: 1\} & 305.0 (9.40\%) \\
     \midrule
     \textbf{Cough} & Present & \{Unlabeled: 0, Negative: -1, Positive: 1\} & 676 (20.84\%) \\
     & Strength & \{Unlabeled: 0, Weak: -1, Strong: 1\} & 360 (11.10\%) \\
     & Induced & \{Unlabeled: 0, Induced: 1\} & 128 (3.95\%) \\
     \midrule
     \textbf{Breath Sounds} & Normality & \{Unlabeled: 0, Normal: 0, Abnormal: 1\} & 1892 (58.34\%) \\ 
     \midrule
     \textbf{Cuff Leak} & Presence & \{Unlabeled: 0, Absent: 0, Present: 1\} & 152 (4.69\%) \\
    \bottomrule
    \end{tabular}
    \caption{Encodings of RT note features included in EF risk models, and the number of patients for whom our LLM pipeline classified any mention of this feature. **Non-pathological sputum colors include clear, white, and tan; pathological sputum colors include yellow, green, pink, red, and rust.
    }
    \label{tab:extractedfeatures}
\end{table*}

\vspace{-1mm}
\section{RT Note Feature Classification}
\label{sec:rtnotes}
\vspace{-1mm}

\definecolor{sputumc}{RGB}{173,216,230}      
\definecolor{coughc}{RGB}{255,204,153}       
\definecolor{breathc}{RGB}{200,230,201}      
\definecolor{cuffleakc}{RGB}{230,200,250}    

\newcommand{\sputumhl}[1]{{\sethlcolor{sputumc}\hl{#1}}}
\newcommand{\coughhl}[1]{{\sethlcolor{coughc}\hl{#1}}}
\newcommand{\breathhl}[1]{{\sethlcolor{breathc}\hl{#1}}}
\newcommand{\cuffleakhl}[1]{{\sethlcolor{cuffleakc}\hl{#1}}}

\newcommand{\labelcolor}[2]{\colorbox{#1}{\strut #2}}

\begin{figure}[t!]
    \centering
    \begin{minipage}[b]{0.5\textwidth}
        \small
        \textbf{RT note excerpt:} \\
        \noindent \texttt{
        ... No vent changes. 
        \breathhl{Breath sounds clear}. 
        Suctioning for 
        \sputumhl{small amounts} of 
        \sputumhl{thick} 
        \sputumhl{white} 
        \sputumhl{secretions}. 
        \coughhl{Strong cough}, +gag, 
        \cuffleakhl{+cuff leak}. 
        Plan SBT and extubate after MRI\quad
        ETT 7\quad 27 @ the teeth\quad \textasciitilde5\string^\quad
        AMV RR 14, Vt 500, +5 ...}
    \end{minipage}
    ~\\ [-2mm]
    \raggedright
    \begin{minipage}[b]{0.49\textwidth}
        \begin{tcolorbox}[
                colframe=blue!40,
                arc=6pt,
                boxrule=0.5pt,
                left=1pt,
                right=1pt,
                top=2pt,
                bottom=2pt
            ]
        \setlength{\parindent}{0pt} 
        \footnotesize
         \textbf{Labels:} \\ [-3mm]
         \begin{itemize}[leftmargin=0em, itemsep=0pt, topsep=0pt]
            \item[] \sputumhl{\textsc{Sputum}}:
            Present=1, Consistency=1, Quantity=1, Color=0
            \item[] \coughhl{\textsc{Cough}}:
            Present=1, Strength=1, Induced=0
            \item[] \breathhl{\textsc{Breath Sounds}}: 0
            \item[] \cuffleakhl{\textsc{Cuff Leak}}: 1
        \end{itemize}
         
        \end{tcolorbox}
    \end{minipage}
    \caption{Excerpt of a respiratory therapy note and feature labels in our dataset. Spans corresponding to feature values are highlighted. 
    }

    \label{fig:example_notes}%
\end{figure}

Respiratory therapists support clinicians in preventing and treating respiratory diseases. They regularly assess clinical features that would necessitate change in respiratory support and document their findings in free text respiratory therapy (RT) notes. 
Our dataset includes 34,834 RT notes collected before each 
patient's initial extubation, and we hypothesize that features documented in these notes may be valuable predictors for extubation failure. 
Relevant features include cough strength and secretion quantity, which both reflect the ability for patients to clear their airways, and are well-established risk factors for extubation failure \citep{ouanes2012nt, chien2008changes, mekontso2006b, duan2021predictive}. 

We define 15 features related to five categories, specified in \autoref{tab:extractmetrics}, based on a review of literature and pulmonologist input.
We then define a novel feature classification task to classify these possible predictors for EF in RT notes. 
For the EF risk assessment task, we are interested in patient-level qualities, so we prompt LLMs to classify entire RT notes, rather than named entity recognition to identify specific note spans. We investigate few-shot prompting due to the lack of a large-scale labeled dataset.

\paragraph{Feature Annotation}
To support prompt engineering and evaluate classification performance, we manually label feature values for a random sample of 400 total RT notes (200 from the training split and 200 from test). We develop annotations collaboratively over multiple rounds of review. The first and second authors label and come to a consensus on labels for each RT note, and these annotations are reviewed and corrected by two practicing critical care pulmonologists. Any disagreements are discussed and a consensus decision is made. An excerpt from an example labeled note is shown in \autoref{fig:example_notes}. We perform prompt engineering exclusively using 200 RT notes from the training split. We report final performance metrics for the classification tasks over the 200 RT notes from the test split.

\paragraph{Few-shot Prompting} We prompt \textsc{Meta-Llama-3-8B-Instruct} \citep{touvron2023llama} to classify feature values. We sample excerpts from the labeled notes of the training split to represent each value of each feature for use as in-context learning examples \citep{brown2020language}. We design separate prompts for suctioning, sputum, cough, breath sounds, and cuff leak features; and prompt the LLM to respond with predefined answers such as \textit{yes/no} for binary features or explicit categorical labels (e.g., \textit{low/medium/high} for sputum amount). All features may be classified as \emph{Unlabeled}, indicating that no explicit feature value is mentioned in the RT note. Temperature was set to 0.01 for all extractions. Final prompts can be found in the supplemental code. 

\paragraph{RT Note Classification Performance}  The mean tokenized RT note length is 258.26 (min 3, max 3178). Over the test set, Macro-F1 for each variable ranges from 0.53 (for spontaneous cough) to 0.99 (for thick sputum), although most F1 values are above 0.80 (see 
\autoref{tab:extractmetrics}).

\paragraph{Classification Results}
We run our final feature classification pipeline over the entire corpus, on RT notes in the 12 hours proceeding extubation events. We have 2,869 such notes from 2,509 of 3,243 valid encounters in our dataset (distribution of notes per patient available in Appendix \ref{app:rt_note_distribution}). Some features are present at high rates 
(abnormal breath sounds at 58.3\%), while others are rare (absent cough at 1.2\%).
Patients with RT notes classified as having high sputum quantity and weak cough co-occur with EF at higher rates than the overall study population,
indicating potential association between these features and EF outcomes. 

\vspace{-1mm}
\section{EF Prediction Methods}
\label{sec:ef_methods}
\vspace{-1mm}

We train logistic regression (LR) and gradient boosting (GB) models to predict, for each patient, their binary outcome of extubation failure (0=success, 1=failure). 

\paragraph{Structured EHR Predictor Variables} For each patient, we collect the following structured EHR variables (part of the \emph{baseline} input variable set):
IMV episode duration, vitals (e.g. SpO$_2$, heart rate, temperature), labs (e.g. hemoglobin, creatinine, calcium), ventilation data (e.g. AutoPEEP, FiO$_2$, tidal volume), diagnoses at admission time (e.g. acute respiratory failure, myocardial infarction, chronic pulmonary disease), and demographics of age and documented sex. We also compute medication doses adjusted for a patient's body weight (e.g. opioid doses, vasopressors, and propofol), which we include as the \emph{medication} input variable set. Since there are multiple measurements, we use the average value of vital, lab, medication, and ventilation features in the 4-hour window before extubation. Appendix \ref{app:predictors} contains a full list of predictors.

\paragraph{Input Variables} We train predictive models with the following combinations of input variables:

\begin{itemize}[noitemsep, topsep=0pt, leftmargin=10pt]
    \item Baseline (B): structured EHR variables such as length of initial intubation, vitals, labs, ventilation data, admission-time diagnoses, age, and documented sex;
    \item Medication (M): normalized medication doses in the 4 hours before extubation; we include medications that are commonly delivered to patients while on IMV, which potentially have negative effects after extubation;
    \item RT Note (N): features classified in respiratory therapy notes as described in \S\ref{sec:rtnotes}. We omit suctioning and spontaneous cough variables because of lower extraction performance (F1$<$0.8) or high correlation with other RT note features (e.g., suctioning presence is highly correlated with sputum presence ($r=0.808$)). Variable values are encoded as described in \autoref{tab:extractedfeatures}.
\end{itemize}

\paragraph{Model Variants} For LR and GB models, we train and test four model variants: one with only base variables (e.g., LR$_B$), one with base and medication variables (e.g., LR$_{B+M}$), one with base and RT note variables (e.g., LR$_{B+N}$) and one with base, medication, and RT note variables (e.g., LR$_{B+M+N}$). These variants allow us to assess the impacts of including each set of variables not included in prior work. In supplementary analysis, to assess the temporal generalizability of our models we also fit variants of the LR$_{B+M+N}$ models on the subset of patients admitted during 2021-22, and report performance on a test set consisting of patients admitted in 2023 (results in Appendix \ref{app:temporal_gen}).

We also assess whether survival analysis models are more effective for estimating probable time to extubation failure. Rather than modeling the probability of a single binary outcome, these models estimate the relationship between predictor variables and the probability a patient will not experience a particular outcome, in our case EF, at a particular time. We assess whether such models are better at predicting EF by accounting for the temporal information of when specific failures occurred or if a patient was discharged. Specifically, we train Cox Proportional Hazards and gradient boosting survival analysis models, fit over the complete set of base, medication, and note variables. The outcome variable of these models is time from extubation to EF, and we censor patients at discharge time. We binarize the outputs of these models by assessing the estimated probability of survival (i.e, not experiencing extubation failure) after seven days.


\paragraph{Model Evaluation \& Analysis} We report AUROC on the held-out test set as our primary metric. Intubated patients form a diverse cohort both in terms of medical conditions and demographic attributes, so we select AUROC because it reflects both positive and negative class performance
\citep{mcdermott2024closer}. We additionally report AUPRC, Sensitivity (Recall), Specificity, PPV (Precision), NPV, F1, and Accuracy. Where appropriate, the threshold for patients deemed high risk is selected based on the prevalence of EF in the training split. 
We report features with high magnitude coefficients in our LR models (all features are normalized before training). 

We further contextualize the performance of our models with the risk predictions made by the readiness checklist CDSS used at \uwname. The checklist includes 19 yes/no questions; per local guidelines, patients with $\ge$2 positive responses are deemed high risk for EF and managed with enhanced post-extubation monitoring. Appendix \ref{app:checklist_predictors} contains the full list of features in this checklist. 

\paragraph{Varying Inclusion Criteria and EF Definition}
To study the impacts of alternate criteria, we train variants of our best performing model: LR$_{B+M+N}$, while varying the inclusion criteria and EF definition. 

\begin{itemize}[itemsep=0pt, topsep=0pt, leftmargin=10pt]
    \item Minimum IMV duration: we vary the minimum IMV duration from 1 hour to 24 hours, while holding the size of the training split constant. For each model variant, we test on a held out split of 20\% of patients meeting the same criteria as the train set. Test split size varies, though any encounter that is used in the test set for any experiment is never used in the training sets for any experiment.
    \item Failure window: we vary the maximum EF window from within 12 hours of extubation  
    to within 336 hours (14 days). We use constant train/test splits differing only by outcome labels, so EF rate ranges from 8.24\% after 12 hours to 22.72\% after 14 days.
\end{itemize}

\begin{table*}[th!]
    \small
    \centering
    \renewcommand\theadfont{}
    \begin{tabular}{L{19mm}L{12mm}ccccccccc}
    \toprule
    Model \& Features & Cohort & AUROC & AUPRC & \thead[tc]{Sens.\\(Recall)} & Spec. & \thead[tc]{PPV\\(Prec.)} & NPV & F1$_+$ & F1$_-$ & Acc. \\
    \midrule
    Checklist** & & — & — & 0.55 & 0.55 & 0.21 & 0.84 & 0.31 & 0.66 & 0.55 \\
    \midrule
    $LR_{B}$ & All & 0.729 & 0.380 & 0.677 & 0.663 & 0.323 & 0.896 & 0.438 & 0.762 & 0.666 \\
    $LR_{B+M}$ & & 0.733 & 0.388 & 0.694 & \textbf{0.670} & 0.333 & 0.902 & 0.450 & \textbf{0.769} & 0.675 \\
    $LR_{B+N}$ & & 0.749 & 0.392 & \textbf{0.750} & 0.657 & \textbf{0.342} & \textbf{0.917} & \textbf{0.470} & 0.766 & 0.675 \\
    $LR_{B+M+N}$ & & \textbf{0.752} & \textbf{0.399} & 0.734 & 0.665 & \textbf{0.342} & 0.913 & 0.467 & \textbf{0.769} & \textbf{0.678} \\
    \midrule
    $GB_{B}$ & All & 0.671 & 0.308 & 0.597 & 0.665 & 0.297 & 0.874 & 0.397 & 0.755 & 0.652 \\
    $GB_{B+M}$ & & 0.671 & 0.307 & 0.597 & 0.657 & 0.292 & 0.872 & 0.393 & 0.750 & 0.646 \\
    $GB_{B+N}$ & & 0.669 & 0.309 & 0.589 & 0.661 & 0.292 & 0.871 & 0.390 & 0.752 & 0.647 \\
    $GB_{B+M+N}$ & & 0.671 & 0.308 & 0.573 & 0.661 & 0.286 & 0.867 & 0.382 & 0.750 & 0.644 \\
    \midrule
    $LR_{B}$ & Patients & 0.696 & 0.386 & 0.680 & 0.633 & 0.330 & 0.881 & 0.444 & 0.737 & 0.643 \\
    $LR_{B+M}$ & with RT & 0.699 & 0.393 & 0.660 & 0.615 & 0.313 & 0.872 & 0.425 & 0.721 & 0.624 \\
    $LR_{B+N}$ & notes & 0.715 & 0.403 & 0.699 & 0.618 & 0.327 & 0.885 & 0.445 & 0.728 & 0.635 \\
    $LR_{B+M+N}$& & \textbf{0.750} & \textbf{0.405} & \textbf{0.702} & \textbf{0.670} & \textbf{0.336} & \textbf{0.904} & \textbf{0.454} & \textbf{0.770} & \textbf{0.676} \\
    \midrule
    \midrule
    \multicolumn{11}{l}{\textbf{Survival Analysis Models}} \\
    $CPH_{B+M+N}$ & All & 0.742 & 0.394 & 0.710 & 0.649 & 0.325 & 0.904 & 0.446 & 0.756 & 0.661 \\
    $GBS_{B+M+N}$ & & 0.677 & 0.336 & 0.677 & 0.565 & 0.270 & 0.881 & 0.386 & 0.689 & 0.587 \\
    \bottomrule
    \end{tabular}
    \caption{Logistic regression (LR) and Gradient boosting (GB) EF prediction models with different input features (B: structured EHR features; M: medications; N: features from RT notes) for all patients and only patients with RT notes. We report results for survival analysis methods (Cox Proportional Hazard (CPH) and Gradient Boosting survival analysis models (GBS) when using all input features. \textbf{Bold} indicates best metric value over that cohort; all cohorts defined as patients with 24-hour minimum intubation time. Thresholded metrics are computed at the proportion of positives in the training split (0.201). \\ [0.5mm]
    \small{**We also show metrics from the existing checklist CDSS to contextualize our results, though model and checklist metrics are not directly comparable since high risk as classified by the checklist is used to provision additional treatment to reduce EF risk.}
    }
    \label{tab:logitmetrics}
\end{table*}

\paragraph{Implementation} We implement all LR and GB models using \texttt{scikit-Learn} \citep{scikit-learn}. We trained survival analysis models using the \texttt{scikit-survival} library \citep{sksurv}. We perform cross-validation over the train split to find hyperparameters maximizing AUROC. 
Final results are reported on the held-out test set. 
Further details in Appendix \ref{sec:implementationdetails}.

\vspace{-1mm}
\section{EF Prediction Results} 
\label{sec:results}
\vspace{-1mm}
\paragraph{RT note variables improve EF prediction} 
Logistic regression models incorporating RT note features demonstrate the strongest performance on our cohort, with consistent gains across most metrics we report (Table~\ref{tab:logitmetrics}). For example, adding RT note features to the baseline logistic regression model increases AUROC from 0.729 (LR$_{B}$; 95\% CI: [0.684, 0.775]) to 0.749 (LR$_{B+N}$; 95\% CI: [0.701, 0.795]). Likewise, adding note variables to the model including base and medication inputs increased performance from 0.733 (LR$_{B+M}$; 95\% CI: [0.684, 0.783]) to 0.752 (LR$_{B+M+N}$; 95\% CI: [0.703, 0.794]). The performance gain when adding RT note variables is even greater among the subset of patients with RT notes within 12 hours of extubation. Appendix \ref{app:demographics} describes performance within demographic subgroups; while metric values varied, no performance differences across sex, age, and racial/ethnic subgroups were found to be statistically significant.


Prior EF prediction research tended to find that GB models outperformed LR models \citep{chen2019prediction, fleuren2021predictors, otaguro2021machine, zhao2021development}. We hypothesize that performance differences may be related to overfitting; to assess this hypothesis we measure model performance over the training set. Over our \emph{training} data, each GB model performed better than the LR model trained using the same inputs. For example, the LR model including all variables attained an AUROC of 0.712 over the training set, whereas the similar GB model attained an AUROC of 0.782 over the training set, indicating that the more complex GB models may not be as robust to distribution shifts between the train and test sets as the LR models.

The Cox Proportional Hazards model fit over all variables attained an AUROC of 0.742  (95\% CI: [0.691, 0.790]). The gradient boosting survival analysis model fit over all variables attained an AUROC of 0.677 (95\% CI: [0.625, 0.732]). These results suggest that for the binary EF prediction task, naive application of survival analysis models may not generate additional predictive performance.

IMV duration is consistently the most important feature in each model (a LR model with only IMV duration as input feature attains an AUROC of 0.658). Other important variables in the LR model by coefficient magnitude include ventilator plateau pressure, SpO$_2$, urea nitrogen, and presence of diffuse traumatic brain injury (ICD-10 code S06.2). Sputum quantity and thickness as classified in RT notes are also among the 20 most important predictors.

Our model demonstrates improvement over the existing checklist CDSS. 
A total of 81 patients in our test set 
assessed 
using the CDSS experienced EF, yet the CDSS classifies only 40 of these patients as high risk.
Our LR$_{B+M+N}$ model correctly classifies 59 of these 81 patients as high risk.
On the other hand, 32 of 40 patients assessed to be high risk by the CDSS are found to be high risk by our LR model, so the two models remain complementary. 
The CDSS attains 21\% precision over patients who experienced EF, while our best model achieves 34\%; although some proportion of patients predicted to be high risk by the CDSS may have avoided EF because clinicians acted upon these high risk assessments.

We conduct calibration analysis comparing logistic regression models with and without RT note features (Appendix \ref{app:calibration}). While neither the LR$_{B+M}$ nor the LR$_{B+M+N}$ models are perfectly calibrated (both have calibration curve slopes over 1), they exhibit similar calibration performance.

\begin{figure*}[ht]
\floatconts
      {fig:inclusionexps}
      {\vspace{-5mm}\caption{Change in AUROC when varying minimum IMV duration and EF window. AUROC tends to decrease as the minimum IMV duration increases (left); AUROC is unchanged as the maximum EF window increases (right). In both experiments, the proportion of patients who experienced EF increases: from 11.26\% to 19.95\% in the case of inclusion variation, and from 8.24\% to 22.72\% in the case of EF definition variation. Each different minimum IMV duration model is trained over varying sets of patients with fixed training split size; training data for each EF window model is identical.\vspace{-3mm}
      }}
      {%
        \subfigure[AUROC varying minimum IMV length]{\label{fig:inclusion_auroc_lr}%
          \includegraphics[width=0.45\linewidth]{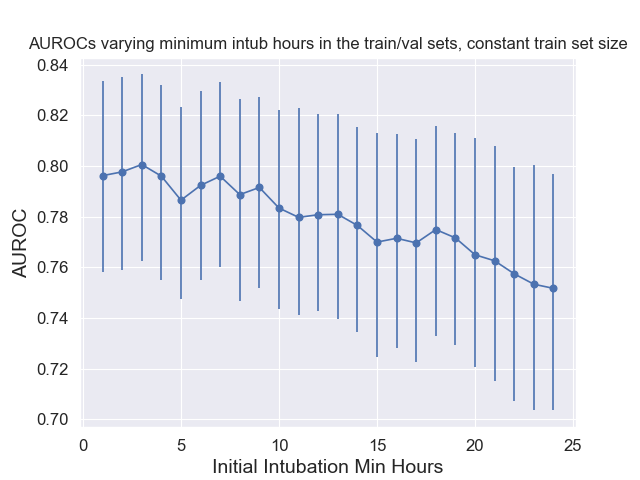}}%
        \qquad
        \subfigure[AUROC varying extubation failure cutoff]{\label{fig:inclusion_f1_lr}%
          \includegraphics[width=0.45\linewidth]{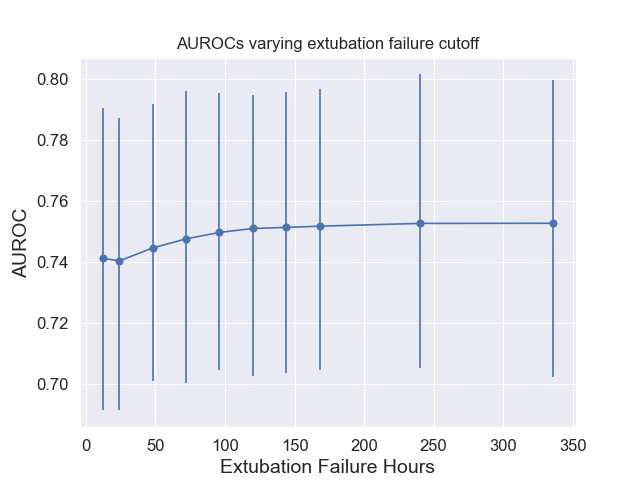}}
      }
\end{figure*}


\paragraph{Increasing minimum IMV duration induces performance trade-offs} 
Model performance 
varies systematically with inclusion criteria. We observe that including patients with shorter duration IMV when training results in a model that more easily classifies positive patients with longer IMV and negative patients with shorter IMV: the model with a 1-hour minimum IMV duration attains recall of 0.258 and specificity of 0.931 for patients whose IMV duration was $<$24h, as compared with recall of 0.855 and specificity of 0.489 for those whose IMV was $\geq$24h.
Correspondingly, AUROC drops from 0.796$\rightarrow$0.752 when varying minimum IMV duration from 1 to 24 hours (Figure~\ref{fig:inclusion_auroc_lr}). 
Positive class F1 increases (0.390$\rightarrow$0.467),
driven by higher precision (0.265$\rightarrow$0.342) with stable overall recall (0.735$\rightarrow$0.734),
while negative class F1 decreases (0.795$\rightarrow$0.777).\footnote{In our experiments, the most permissive minimum IMV duration (with same train set size) results in highest AUROC ($\approx$0.80), similar to metrics reported in prior work.} 
GB models exhibited similar AUROC and positive F1 trends, but did not consistently vary in negative class F1 (Appendix \ref{app:more_inclusion_criteria}).


Changes in predictor importance between the 1-hour and 24-hour minimum IMV duration cohorts indicate possible cohort-level differences. While 
IMV episode length, SpO$_2$, urea nitrogen, plateau pressure, and hemoglobin are the top four predictors in each model, 
Acute Respiratory Failure, age$>$60, and FiO$_2$ are among the top predictors for the 24-hour minimum model, but not the 1-hour minimum model
(Appendix \autoref{tab:logitcoeffs}). Additionally, characteristics of these groups vary: for instance, 12.7\% of patients who were intubated $<$24 hours were in a surgical unit for their entire intubation, as compared with 5.2\% of patients who were intubated $\geq$24 hours (p$<$0.001).



\paragraph{Varying EF window does not systematically alter metrics} When increasing maximum EF window from 12 to 336 hours, we do not observe major changes in overall predictor performance per AUROC (Figure~\ref{fig:inclusion_f1_lr}) or positive/negative class F1 (Appendix \autoref{fig:cutoffexps} shows changes in F1), despite the increase in EF prevalence from 8\% to $>$21\%. IMV duration remains the most important predictive feature across different choices for EF window. 

\vspace{-1mm}
\section{Discussion \& Conclusion}
\vspace{-1mm}
Our pipeline classifies a novel set of variables in free-text respiratory therapy notes, a type of clinical note that as far as we know, has not been studied in prior clinical LLM work. 
LLMs demonstrate strong performance in classifying these features, especially those related to sputum, breath sounds, cuff leak presence, and cough presence and strength. 

Predictive models including RT note features perform better than models without these features. These increases remain consistent to robustness checks such as inclusion of medication variables in the model, and disappear when shuffling the columns corresponding to the RT note features. These results imply clinically relevant improvements in predictive performance: \citet{krinsley2012optimal} suggest 5\% may be a reasonable target extubation failure rate (i.e, only 5\% of the population are false negatives for detection of extubation failure). To attain this extubation failure rate over our test data, a model must have sensitivity of 74.2\%. At this sensitivity, the base LR model has a FPR of 40.0\%, while the LR model augmented with RT note features has a FPR of 34.1\%. This difference in FPR corresponds to 5 fewer false alarms per 100 patients distributed identically to our test set, preventing harmful impacts of unnecessary continuation of IMV. A corresponding decision curve analysis indicates that in settings where clinicians target roughly one true case of extubation failure per five false alarms, the additional RT note features may yield one additional true positive per 100 patients treated. Appendix \ref{app:dca} contains further decision curve analysis details. 

We identify high volume of sputum, thick sputum, and weak cough as associated with higher EF risk, so these features may be valuable to document as structured EHR elements in the future. Meanwhile, other significant features in our model, such as long IMV duration, low SpO$_2$, blood urea nitrogen, and brain injury are known to be associated with extubation failure, providing external validity for our results \citep{igarashi2022machine, vidotto2008prediction}. 

While prior work reports that gradient boosting variants outperformed logistic regression for EF prediction, our work demonstrates that this may not apply to every EF prediction setting. As \citet{Christodoulou2019ASR} show, other machine learning methods do not exhibit significant improvements over logistic regression for clinical prediction models, so there is precedent for such findings. We additionally find that models developed using survival analysis objectives did not exhibit significantly improved performance on binary extubation failure prediction tasks.

Models trained and assessed using different inclusion criteria exhibit performance differences, indicating a potential threat to model generalizability. 
In our results, AUROC decreases and F1 increases as the minimum IMV duration threshold is raised. Higher IMV duration is associated with higher EF risk, and setting a higher threshold tends to remove more-easily classified lower-risk patients from the population, leading to lower AUROC \citep{mcdermott2024closer}. 
The relatively poor performance of models greater minimum intubation durations  reflect the difficulty of distinguishing risk among patients where the task is clinically meaningful: clinicians are less likely to need risk assessment support for patients with shorter, post-operative intubations.
This highlights a broader implication: collaboration with clinicians is essential to ensure that model tasks can be specified in relation to meaningful clinical needs.

Our analysis also yields some 
counterintuitive observations, such as diagnosis of acute respiratory failure (ARF) 
being associated with \textit{lower} odds of EF. 
It is possible that ARF may not be documented as a diagnosis when comorbid with other IMV causes such as head injury or stroke.
These diagnoses may induce treatment differences, possibly altering EF risk.
Due to ours and similar studies being retrospective, without counterfactuals, it is difficult to disentangle these 
mechanisms. We leave work that controls for diagnosis and treatment effects to future clinical studies.

In this work, a key advantage of using classified feature values versus dense note embeddings 
is interpretability in the downstream model (at the cost of less expressivity). Future work could investigate how best to preserve additional details from RT notes and other unstructured notes in downstream models. 
Another future direction is investigating how alternate modeling methods, such as time series models, may improve EF prediction, especially when incorporating RT note features.
Lastly, EF risk factors may also be associated with demographic groups; e.g., \citet{thille2023sex} uncover sex-related differences in IMV and EF. Future work should investigate whether the features we uncover are broadly useful or are only associated with increased EF risk in specific demographic or clinical groups.

\paragraph{Limitations} 
Due to dataset constraints, we are limited to studying EF in a single healthcare system located in a large US city. No comparable dataset is publicly available (other public EHR datasets do not contain RT notes), impeding external validation of our results. Both our LLM feature classification pipeline and the downstream EF prediction models may fail to generalize to other clinical notes and patient populations, other languages besides English, or even to the same hospitals over time as the population of intubated patients can change.

We lack treatment counterfactuals: outcomes may have differed had patients not been treated as high risk in the clinical setting, so we cannot assess causal associations between patient features and extubation outcomes. The decision to extubate can also be influenced by non-clinical factors such as family wishes or transitions to palliative care. While DNR/DNI are the most objective way of excluding patients ineligible for re-intubation, future work should investigate whether relevant non-clinical factors can be derived from unstructured notes and quantify their impacts on extubation/re-intubation decisions and outcomes. A larger sample of manually labeled notes may be useful for future work aiming to develop a more robust pipeline.

\paragraph{Conclusion}
We demonstrate that LLMs in few-shot settings can
classify features relevant to extubation failure in free text clinical notes collected during IMV.
These features improve performance of downstream EF risk models, enabling identification of additional risk factors that may be useful to include in future CDSS. Furthermore, we describe how inconsistent cohort inclusion criteria are prevalent in related work yet drive changes in model performance, demonstrating a threat to generalizability beyond differences across hospital settings. This result reveals the need for standardized task definitions to enable model comparability and support translation of risk prediction models into clinical practice.

\section*{Author Contributions}
The first author led data extraction, conducting experiments, analysis of results, and writing this manuscript. The second author aided in designing the LLM note labeling pipeline, labeling clinical notes, and contributing to writing in Section 4. The third and fourth authors are practicing intensive care pulmonologists at \uwname; they provided assistance in data collection and interpretation, developing clinical note classification schema, final labeling of clinical notes, selection of experiments, interpretation of experiment results, and clinical motivation and background included in this manuscript. The final author provided detailed guidance on all aspects of this research. 

\section*{Acknowledgements}

This work was supported by the University of Washington Institute of Medical Data Science Pilot Award, the University of Washington eScience Institute's Cloud Credits for Research and Teaching Program, and gift funds from the Allen Institute for AI. The first author was partially supported by the National Science Foundation CSGrad4US Fellowship.


\bibliography{main}


\appendix



\section{RT Note Distribution}
\label{app:rt_note_distribution}
\begin{figure}[t!]
    \centering
    \includegraphics[width=\linewidth]{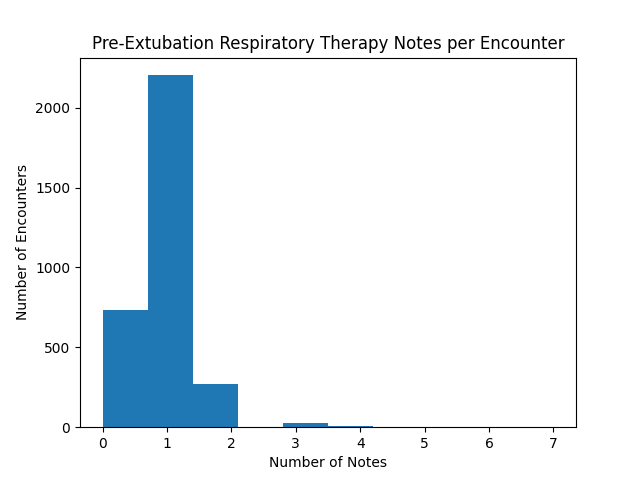}%
    \caption{Number of respiratory therapy notes collected within 12 hours prior to extubation}%
    \label{fig:numnoteshist}%
\end{figure}

\begin{table*}[t!]
\small
    \centering
    \begin{tabular}{ll|ll|ll}
    \toprule
     & & \multicolumn{2}{c}{\textbf{24h Minimum Intubation}} & \multicolumn{2}{c}{\textbf{1h Minimum Intubation}} \\
    \midrule
    \textbf{Factor} & \textbf{Group} & N Patients (\%) & N Failed (\%) & N Patients (\%) & N Failed (\%) \\
    \midrule
    \textbf{Gender} & Non-Male & 1045 (32.22\%) & 202 (19.33\%) & 2270 (32.77\%) & 241 (10.62\%) \\
     & Male & 2198 (67.78\%) & 445 (20.25\%) & 4658 (67.23\%) & 539 (11.57\%) \\
     \hline
     \textbf{Age} & $\leq$60 & 2355 (72.62\%) & 436 (18.51\%) & 4828 (69.69\%) & 517 (10.71\%) \\
     & $>$60 & 888 (27.38\%) & 211 (23.76\%) & 2100 (30.31\%) & 263 (12.52\%) \\
     \hline

    \textbf{Race/Ethnicity}$^\dagger$ & Asian & 225 (6.94\%) & 54 (24.00\%) & 477 (6.89\%) & 64 (13.42\%) \\
     & Black & 338 (10.42\%) & 61 (18.05\%) & 619 (8.93\%) & 72 (11.63\%) \\
     & Hispanic & 309 (9.53\%) & 51 (16.50\%) & 606 (8.75\%) & 65 (10.73\%) \\
     & White & 2009 (61.95\%) & 414 (20.61\%) & 4513 (65.14\%) & 501 (11.10\%) \\
     & Other & 362 (11.16\%) & 67 (18.51\%) & 713 (10.29\%) & 78 (10.94\%) \\
     \midrule
     \textbf{Total} & - & 3243 (100\%) & 647 (19.95\%) & 6928 (100\%) & 780 (11.26\%) \\
    \bottomrule
    \end{tabular}
    \caption{Patient demographics. Extubation failure rates are computed with a maximum failure window of 7 days. $^\dagger$The Hispanic group includes Hispanic patients of any race and other groups include non-Hispanic patients of any race. 
    } 
    \label{tab:demographics}
\end{table*}
Most encounters had at least one respiratory therapy note collected in the 12 hours prior to extubation. \autoref{fig:numnoteshist} displays number of notes per encounter in the primary cohort of 3,243 encounters.

\section{RT Note Extraction Confusion Matrices}
\autoref{fig:confmtx_cough} contains confusion matrices for the cough features extracted.

\begin{figure*}[htbp]
\floatconts
      {fig:confmtx_cough}
      {\caption{Cough feature confusion matrices for the LLM entity extraction pipeline}}
      {%
        \subfigure[Cough Present Confusion Matrix]{\label{fig:cough_pres_mtx}%
          \includegraphics[width=0.45\linewidth]{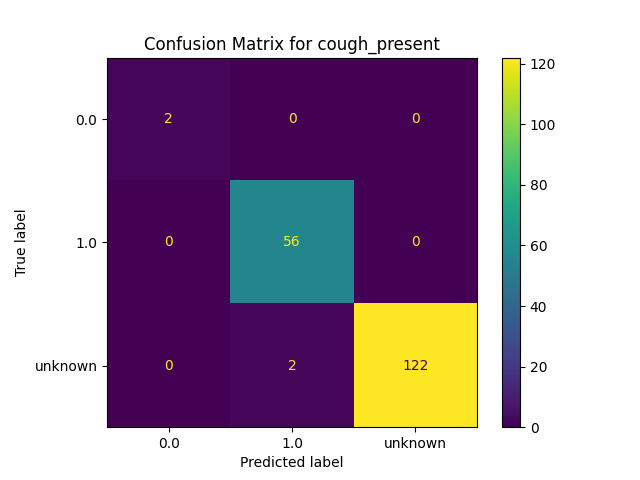}}
        \qquad
        \subfigure[Cough Induced Confusion Matrix]{\label{fig:cough_induced_mtx}%
          \includegraphics[width=0.45\linewidth]{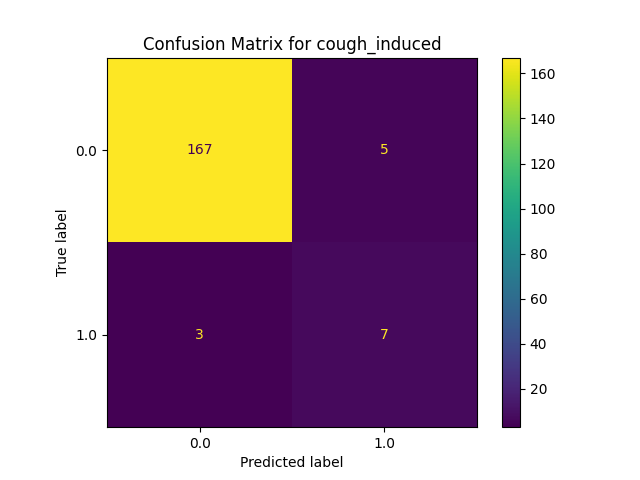}}
        \subfigure[Cough Spontaneous Confusion Matrix]{\label{fig:cough_spont_mtx}%
          \includegraphics[width=0.45\linewidth]{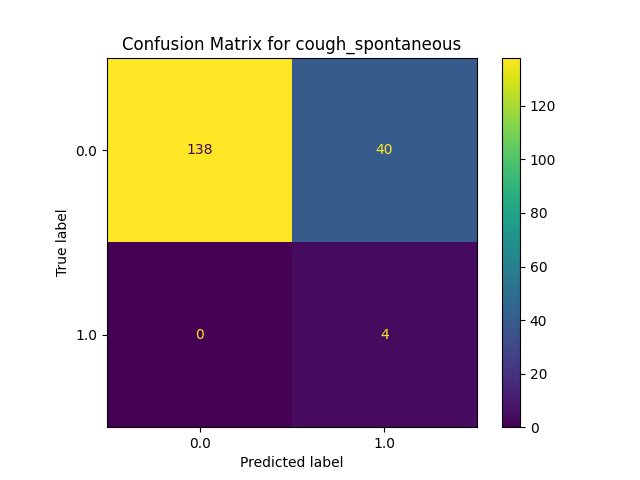}}
        \qquad
        \subfigure[Cough Strong Confusion Matrix]{\label{fig:cough_strong_mtx}%
          \includegraphics[width=0.45\linewidth]{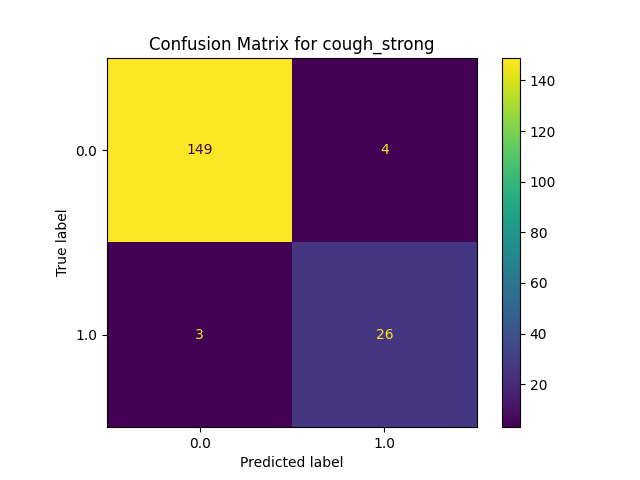}}
        \subfigure[Cough Weak Confusion Matrix]{\label{fig:cough_weak_mtx}%
          \includegraphics[width=0.45\linewidth]{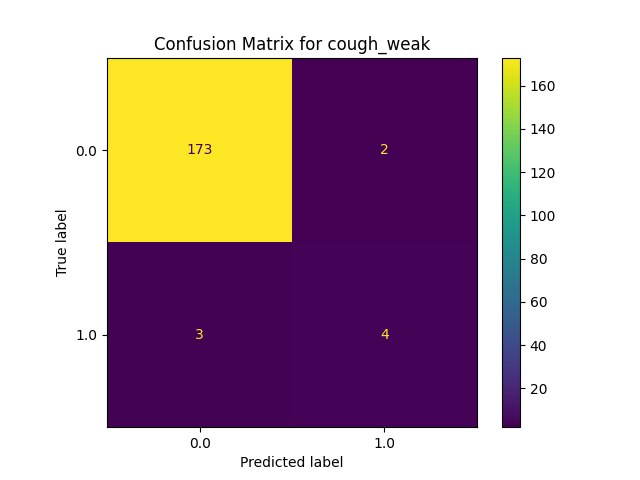}}

      }
\end{figure*}
\begin{figure*}[htbp]
\floatconts
      {fig:confmtx_sputum}
      {\caption{Sputum feature confusion matrices for the LLM entity extraction pipeline}}
      {%
        \subfigure[Sputum Present Confusion Matrix]{\label{fig:sputm_present_mtx}%
          \includegraphics[width=0.45\linewidth]{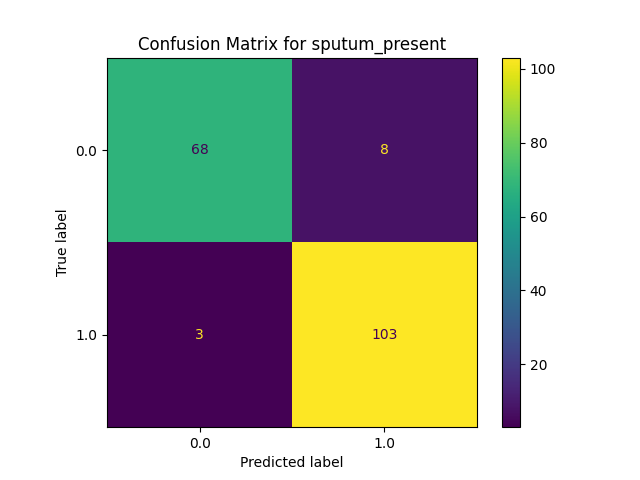}}
        \qquad
        \subfigure[Sputum Quantity Confusion Matrix]{\label{fig:sputum_quant_mtx}%
          \includegraphics[width=0.45\linewidth]{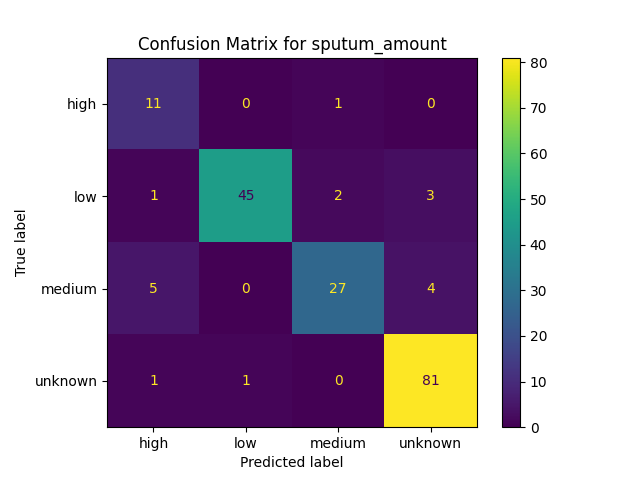}}
        \subfigure[Sputum Color Confusion Matrix]{\label{fig:sputum_color_mtx}%
          \includegraphics[width=0.45\linewidth]{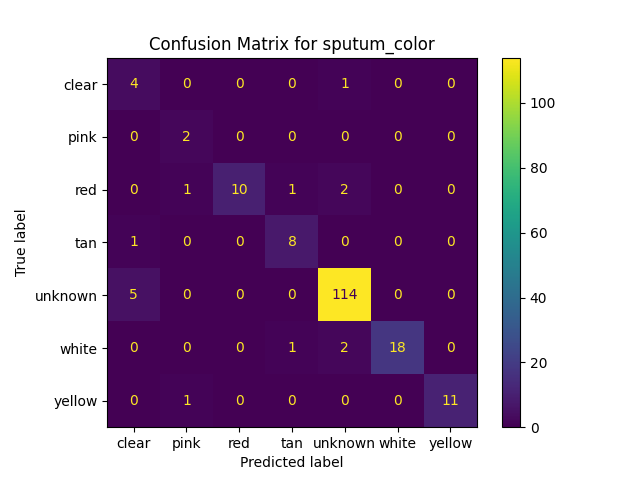}}
        \qquad
        \subfigure[Sputum Thick Confusion Matrix]{\label{fig:sputum_thick_mtx}%
          \includegraphics[width=0.45\linewidth]{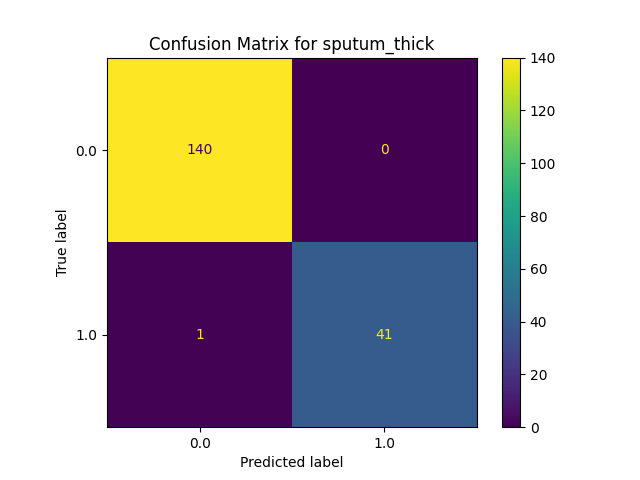}}
        \subfigure[Sputum Thin Confusion Matrix]{\label{fig:sputum_thin_mtx}%
          \includegraphics[width=0.45\linewidth]{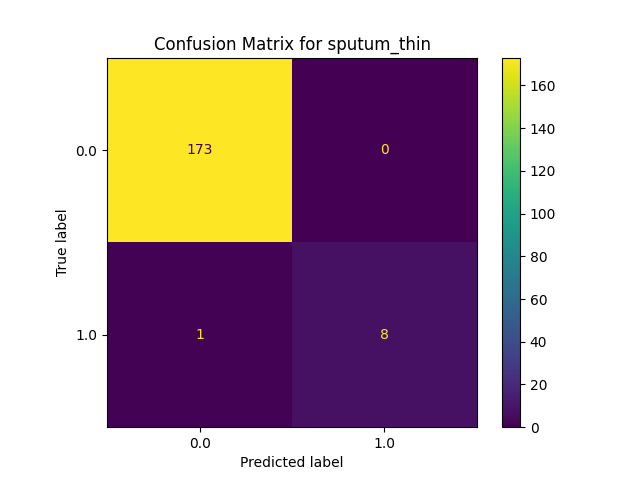}}

      }
\end{figure*}

\begin{figure*}[htbp]
\floatconts
      {fig:confmtx_other}
      {\caption{Breath Sounds, Cuff Leak, and Suctioning feature confusion matrices for the LLM entity extraction pipeline}}
      {%
        \subfigure[Breath Sounds Confusion Matrix]{\label{fig:breathsound_mtx}%
          \includegraphics[width=0.45\linewidth]{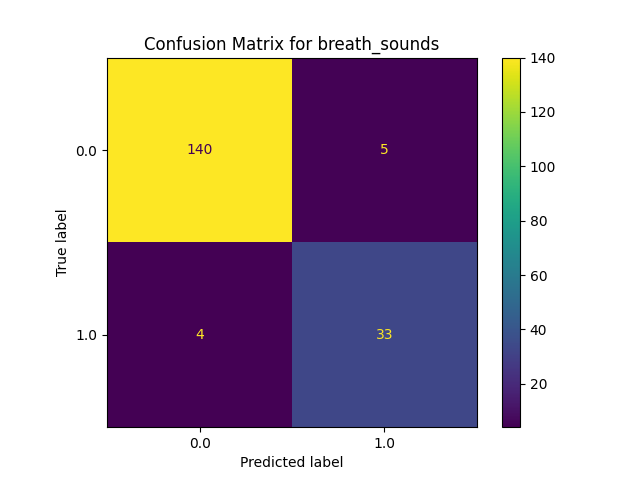}}
        \qquad
        \subfigure[Cuff Leak Confusion Matrix]{\label{fig:cuff_leak_mtx}%
          \includegraphics[width=0.45\linewidth]{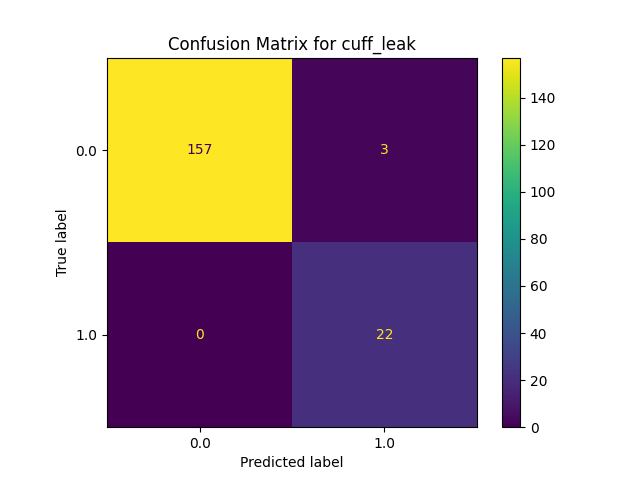}}
        \subfigure[Suctioning Present Confusion Matrix]{\label{fig:suctioning_pres_mtx}%
          \includegraphics[width=0.45\linewidth]{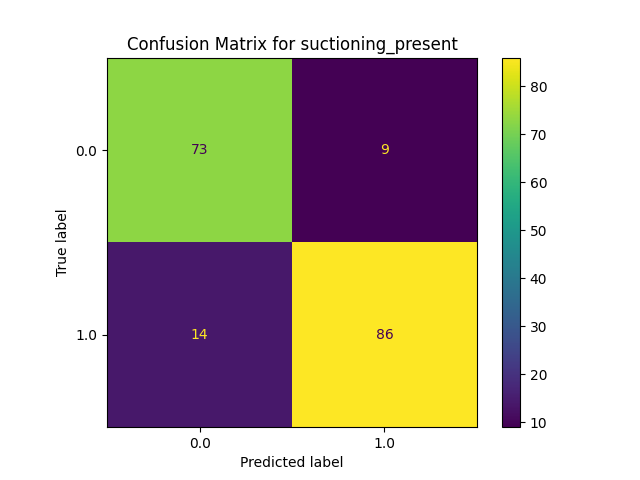}}
        \qquad
        \subfigure[Suctioning Oral Confusion Matrix]{\label{fig:suctioning_oral_mtx}%
          \includegraphics[width=0.45\linewidth]{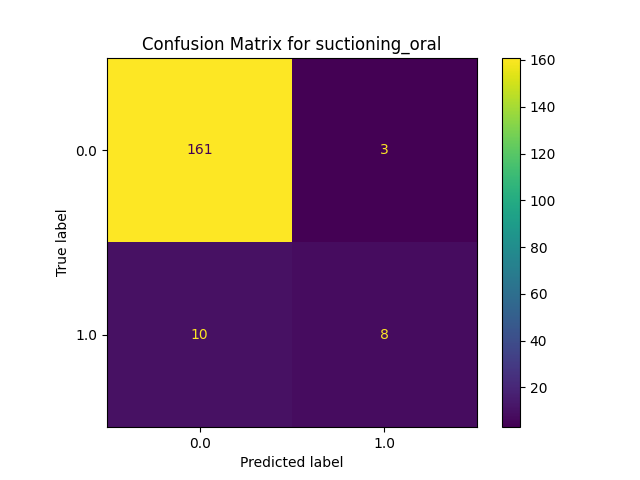}}
        \subfigure[Suctioning Endotracheal Confusion Matrix]{\label{fig:suction_ett_mtx}%
          \includegraphics[width=0.45\linewidth]{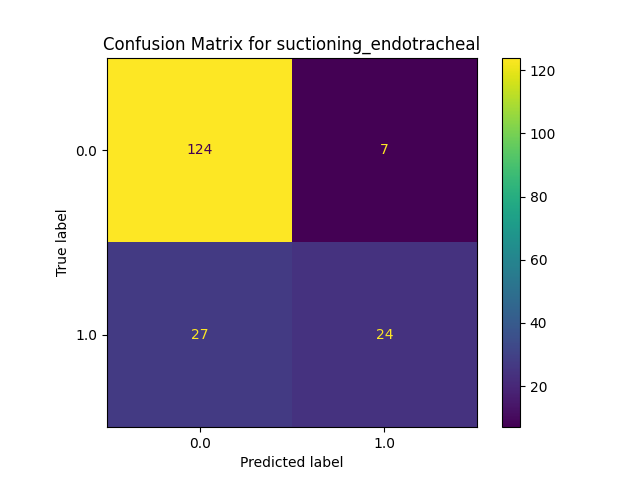}}

      }
\end{figure*}

\section{Checklist CDSS Features}
\label{app:checklist_predictors}
The current readiness exam checklist at our institution contains the following predictor variables, of which any two being met indicates that a patient is at high risk for EF:
\begin{enumerate}[noitemsep, topsep=0pt, leftmargin=18pt]
    \item History of difficult airway
    \item Restricted airway access
    \item Concern over reintubation
    \item C-spine surgery $>$ 3 levels with operative time $>$ 5h or blood loss $>$ 300 mL
    \item Posterior fossa pathology
    \item BMI greater than or equal to 40 kg/m$^2$
    \item Lack of cuff leak
    \item Lack of spontaneous cough
    \item Tracheal suctioning frequency more than twice per hour
    \item Frequent oral suctioning
    \item Failed more than 3 previous SBTs
    \item Age over 60
    \item Male gender
    \item Coma
    \item Chronic lung disease
    \item Positive cardiac history
    \item End stage kidney disease
\end{enumerate}
 

\section{Model Implementation Details}
\label{sec:implementationdetails}
We compute the following medication dosage received by each patient: oral morphine milligram equivalent opioids \citep{broglio2022approximate}, diazepam-equivalent benzodiazepine \citep{ashton2002benzodiazepines}, propofol-equivalent vasopressors \citep{goradia2021vasopressor}, crystalloid volume, and a binary variable indicating whether a patient received a neuromuscular blockade.

All input features are normalized to have mean of 0 and unit standard deviation, and missing features are imputed to the training set mean. We measured model performance metrics averaged over 8-fold cross validation of the train set, and we perform a grid search over number of estimators, learning rate, minimum split samples, minimum leaf samples, and tree depth to find optimal hyperparameters for the gradient boosting model, and maximum iterations and L2 regularization strength \textit{C} for the logistic regression models. We then fit a model using the optimal hyperparameters over all encounters not reserved for testing. We also train and test a model only including RT note features as inputs, which achieves AUROC 0.605 (95\% CI: [0.548, 0.658]), indicating predictive value beyond chance.

\section{Predictors for LR Models}
\label{app:predictors}

\begin{table*}[t!]
    \small
    \centering
    \begin{tabular}{L{43mm}C{23mm}C{30mm}C{33mm}C{15mm}}
    \toprule
    Predictor & Mean (std) & Failed Mean (std) & Non-Failed Mean (std) & P-Value \\
    \midrule
    Initial IMV episode length & 93.52 (95.57) & 127.41 (117.10) & 84.98 (87.30) & $\ll$0.001 \\
    Plateau pressure & 17.95 (3.72) & 18.98 (4.11) & 17.69 (3.57) & $\ll$0.001 \\
    SpO2 & 96.90 (1.94) & 96.45 (2.12) & 97.01 (1.88) & $\ll$0.001 \\
    Urea nitrogen & 29.12 (22.47) & 35.13 (26.80) & 27.62 (21.00) & $\ll$0.001 \\
    Tempurature & 98.60 (1.24) & 98.79 (1.27) & 98.55 (1.22) & $<$0.001 \\
    Respiration & 17.78 (4.16) & 18.67 (4.28) & 17.56 (4.10) & $\ll$0.001 \\
    Number of SBTs & 1.39 (1.75) & 1.77 (2.02) & 1.29 (1.66) & $\ll$0.001 \\
    Hemoglobin & 9.71 (2.04) & 9.34 (1.92) & 9.80 (2.06) & $<$0.001 \\
    Sputum amount & 0.71 (0.91) & 0.85 (0.99) & 0.67 (0.89) & 0.014 \\
    FiO2 (\%) & 31.33 (9.33) & 32.82 (10.12) & 30.94 (9.08) & 0.003 \\
    \toprule
    Predictor & Count (\%) & Failed Count (\%) & Non-Failed Count (\%) & P-Value \\
    \midrule
    Diffuse traumatic brain injury & 55 (2\%) & 22 (4\%) & 33 (2\%) & $<$0.001 \\
    Acute respiratory failure & 264 (10\%) & 32 (6\%) & 232 (11\%) & $<$0.001 \\
    Cerebralvascular disease & 344 (13\%) & 90 (17\%) & 254 (12\%) & 0.004 \\
    Hemiplegia or paraplegia & 221 (9\%) & 62 (12\%) & 159 (8\%) & 0.003 \\
    Age $>$ 60 & 776 (30\%) & 185 (35\%) & 591 (28\%) & 0.003 \\
    \bottomrule
    \end{tabular}
    \caption{Summary statistics for most important predictors by feature coefficient in LR$_{B+M+N}$ model.}
    \label{tab:sumstats}
\end{table*}

\begin{table*}[ht]
    \small
    \centering
    
    \begin{tabular}{lc|lc}
    \toprule
         \multicolumn{2}{c}{\textbf{24h Minimum Intubation}} & \multicolumn{2}{c}{\textbf{1h Minimum Intubation}} \\
        Predictor & Coefficient & Predictor & Coefficient \\
        \midrule
Initial IMV Episode Length & 0.0955 & Initial IMV Episode Length & 0.1756 \\
Plateau Pressure (cm H2O) & 0.0679 & SpO2 & -0.1313 \\
SpO2 & -0.0664 & Urea Nitrogen & 0.1051 \\
Urea Nitrogen & 0.0601 & Plateau Pressure (cm H2O) & 0.0815 \\
Diffuse traumatic brain injury & 0.0541 & Hemoglobin & -0.0721 \\
Temperature & 0.0475 & Respiratory Rate & 0.0720 \\
Respiratory Rate & 0.0466 & Temperature & 0.0687\\
N SBTs Before Extubation & 0.0459 & N SBTs Before Extubation & 0.0670 \\
Hemoglobin & -0.0434 & Sputum amount & 0.0620 \\
\textbf{Acute Resp Failure} & -0.0413 & \textbf{Calcium} & -0.0619 \\
Cerebrovascular disease & 0.0410 & Cerebrovascular disease & 0.0610 \\
Hemiplegia or paraplegia & 0.0398 & Diffuse traumatic brain injury & 0.0588 \\
Sputum amount & 0.0392 & \textbf{Propofol in last 4h} & -0.0522 \\
\textbf{Age $>$ 60} & 0.0392 & Sputum thickness & 0.0475 \\
\textbf{FiO$_2$} (\%) & 0.0372 & \textbf{Heart rate} & 0.0412 \\
\bottomrule
    \end{tabular}
    \caption{Variables with the 15 highest magnitude coefficients in the logistic regression BMN model (all features normalized to zero mean and unit standard deviation); note that sputum amount associated with higher risk of extubation failure. Bolded variables are not among the top 20 most important features for the other model. Among all listed features, only FiO$_2$ has different sign across models.}
    \label{tab:logitcoeffs}
\end{table*}

The four most important coefficients were common between the LR model trained with a 24 hour minimum IMV duration and a 1 hour minimum duration. These features also are present in prior literature indicating potential predictors for extubation failure. However, several of the less important features are not shared between models, and the sign on the coefficient for FiO$_2$ measurements differs between models. \autoref{tab:logitcoeffs} contains a set of the most important coefficients. Sputum amount consistently has the highest magnitude coefficient among the features derived from the RT notes. \autoref{tab:sumstats} shows summary stats for this cohort.

Below is the full list of variables included in the $LR_{B+M+N}$ model: \begin{itemize}[noitemsep, topsep=0pt, leftmargin=10pt]
    \item Vitals: mean arterial pressure, heart rate, respiratory rate, SpO$_2$, and temperature. 
    \item Labs: Anion Gap, Calcium, CO$_2$ (total), Chlorine, Creatinine, Glucose, Hemoglobin, MCV, Platelet Count, Potassium, Sodium, Urea Nitrogen, White Blood Cell Count, Arterial pH, C-Reactive Protein
    \item Ventilation Data: Duration of IMV, Number of Spontaneous Breathing Trials, Fio$_2$, Insp. Flow, Minute Ventilation, Plateau Pressure, Insp. Pressure, AutoPEEP, Observed Tidal Volume
    \item Diagnoses: Mycocardial Infarction, Congestive Heart Failure, Peripheral Vascular Disease, Cerebrovascular Disease, Dementia, Chronic pulmonary disease, Rheumatic disease, Peptic ulcer disease, Mild liver disease, Diabetes without chronic complication, Diabetes with chronic complication, Hemiplegia or paraplegia, Renal disease, Malignancy, Moderate or severe liver disease, AIDS/HIV, COVID-19, Diffuse traumatic brain injury, Spinal cord injury below neck, Septic shock, Atherosclerotic heart disease, Anemia, Acute respiratory failure, Sleep apnea, COVID-19 exposure, Nicotine dependence (cigarettes), GERD, Hypokalemia, Hypo-osmolality and Hyponatremia, Acute kidney failure, Hyperlipidemia, Hypertension
    \item Medications: Opioid dose, Benzodiazepine dose, Vasopressor dose, Crystalloid dose, Propofol dose, Neuromuscular Blockade presence
    \item Demographics: Age over 60, Documented Male Sex
    \item RT Note Variables: Sputum presence, Sputum thickness, Sputum quantity, Pathological sputum color, Cough presence, Cough strength, Induced cough, Cuff leak, Abnormal breath sounds
\end{itemize}
\begin{figure}[t!]
    \centering
    \includegraphics[width=0.8\linewidth]{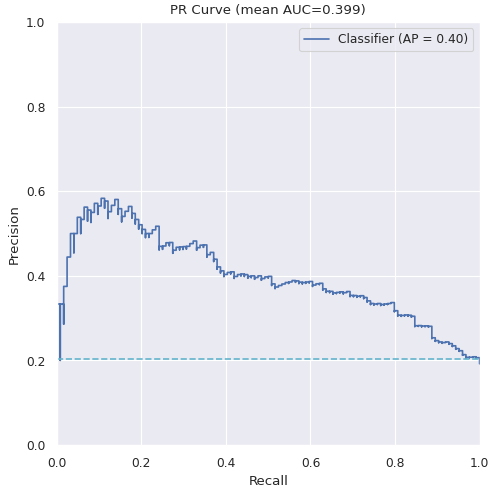}%
    \caption{P/R curve for the $LR_{B+M+N}$ model. This model attained an AUROC of 0.75, but fails to achieve precision greater than 0.6 over the test set. 
    }%
    \label{fig:aurocauprc}%
\end{figure}


\section{Analysis of Inclusion Criteria}
\label{app:more_inclusion_criteria}
\begin{figure*}[htbp]
\floatconts
      {fig:cutoffexps}
      {\caption{Changes in F1 for positive and negative class in inclusion variation and EF definition variation experiments. When changing inclusion criteria, positive class F1 increases (primarily due to increases in precision) whereas negative class F1 decreases. Neither class F1 changes greatly as the maximum hours before valid extubation failure changes.}}
      {%
        \subfigure[Minimum IMV Length Variation; F1]{\label{fig:inclusion_criteria_f1}%
          \includegraphics[width=0.45\linewidth]{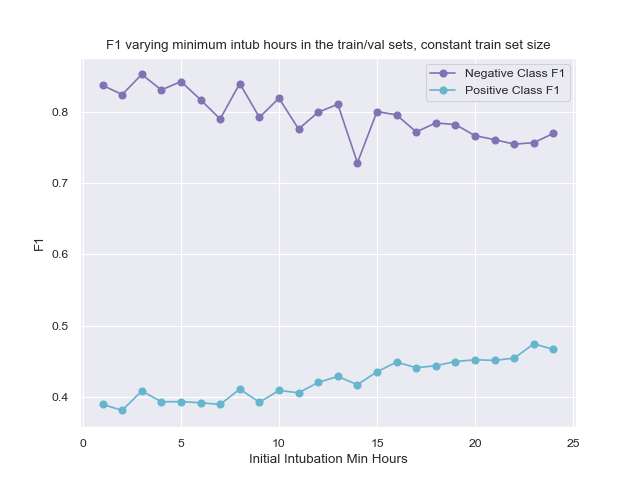}}%
        \qquad
        \subfigure[EF Window Variation; F1]{\label{fig:_failure_window_f1}%
          \includegraphics[width=0.45\linewidth]{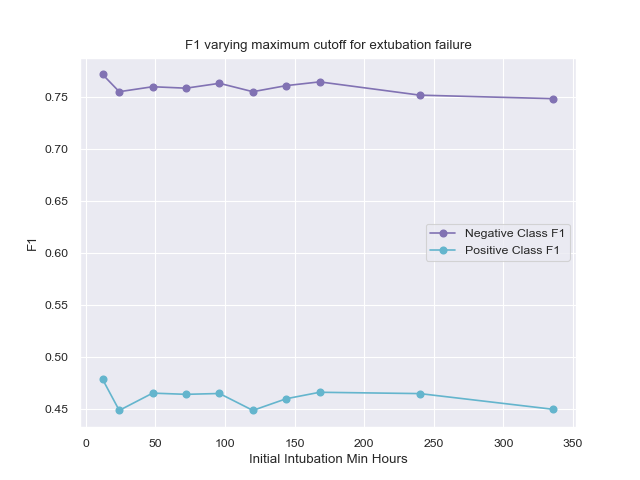}}
      }
\end{figure*}

In addition to AUROC, we also measure F1 scores for the positive (patients who did experience EF) and negative (patients who did not experience EF) examples in the test set. In the same $LR_{B+M+N}$ model, we observe that negative class F1 tends to decrease and positive class F1 tends to increase as minimum IMV duration increases (Figure~\autoref{fig:inclusion_criteria_f1}). However, while a gradient boosting model demonstrates the same trends for AUROC and positive class F1, it does not exhibit the same trend in negative class F1, indicating metric changes may depend on model type (\autoref{fig:gbinclusionexps}). The positive and negative class F1s for the EF Window variation experiments do not vary systematically as EF Window varies (Figure~\autoref{fig:_failure_window_f1}).

\begin{figure*}[h!]
\floatconts
      {fig:gbinclusionexps}
      {\caption{Experiments in varying initial intubation length, fitting \textbf{gradient boosting} models; AUROC and positive class F1 exhibit the same trends as logistic regression models, whereas negative class F1 first decreases, then increases (see \autoref{fig:inclusionexps}). This pattern is largely due to changes in specificity: specificity for the model with a 1 hour minimum IMV duration is 0.683, 0.560 for the 12 hour model, and increases to 0.728 for the 17 hour model (these models' respective NPVs are 0.949, 0.934, and 0.896). Increases in positive class F1 are again largely related to increases in precision.}}
      {%
        \subfigure[Minimum IMV Duration Length Variation; AUROC]{\label{fig:inclusion_auroc}%
          \includegraphics[width=0.45\linewidth]{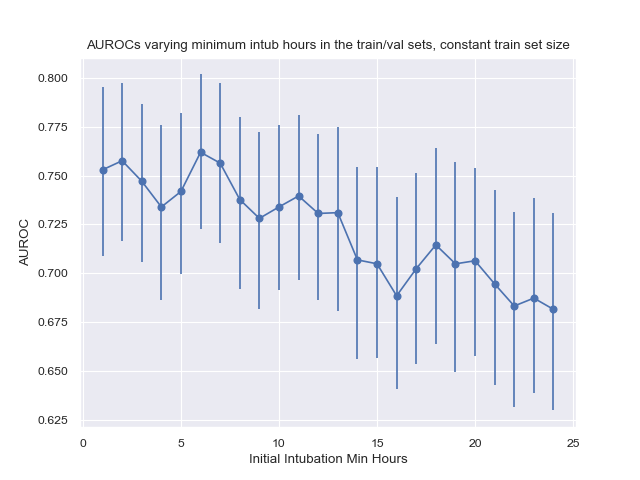}}%
        \qquad
        \subfigure[Minimum IMV Duration Length Variation; F1]{\label{fig:inclusion_f1}%
          \includegraphics[width=0.45\linewidth]{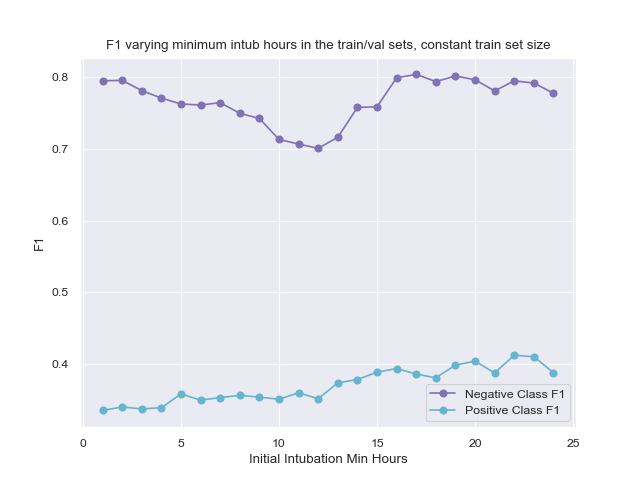}}
      }
\end{figure*}

\begin{table*}
    \small
    \centering
    \renewcommand\theadfont{}
    \begin{tabular}{L{32mm}ccccccccc}
    \toprule
    \thead[tl]{Group} & AUROC & AUPRC & \thead[tc]{Sens.\\(Recall)} & Spec. & \thead[tc]{PPV\\(Prec.)} & NPV & F1$_+$ & F1$_-$ & Acc. \\
    \midrule
        All & 0.752 & 0.399 & 0.734 & 0.665 & 0.342 & 0.913 & 0.467 & 0.769 & 0.678 \\
    \midrule
    Non-Male Patients & 0.713 & 0.374 & 0.676 & 0.671 & 0.325 & 0.898 & 0.439 & 0.768 & 0.672 \\
    Male Patients & 0.765 & 0.415 & 0.759 & 0.662 & 0.349 & 0.920 & 0.478 & 0.770 & 0.681 \\
    \midrule
    Age $\leq$ 60 & 0.749 & 0.401 & 0.714 & 0.681 & 0.335 & 0.914 & 0.456 & 0.781 & 0.687 \\
    Age $>$ 60 & 0.754 & 0.439 & 0.808 & 0.581 & 0.368 & 0.909 & 0.506 & 0.709 & 0.634 \\
    \midrule
    Asian Patients & 0.820 & 0.530 & 0.931 & 0.643 & 0.474 & 0.964 & 0.628 & 0.771 & 0.717 \\
    Black Patients & 0.756 & 0.427 & 0.633 & 0.669 & 0.297 & 0.892 & 0.404 & 0.765 & 0.663 \\
    Hispanic Patients & 0.715 & 0.306 & 0.750 & 0.659 & 0.295 & 0.933 & 0.424 & 0.772 & 0.673 \\
    White Patients & 0.673 & 0.434 & 0.600 & 0.675 & 0.316 & 0.871 & 0.414 & 0.761 & 0.660 \\
    Other Patients & 0.748 & 0.434 & 0.677 & 0.676 & 0.323 & 0.902 & 0.438 & 0.773 & 0.677 \\
    \bottomrule
    \end{tabular}
    \caption{Model performance for the $LR_{B+M+N}$ model stratified by mutually exclusive groups. Patients documented as male, who make up 67.78\% of our dataset, attain better performance, including better precision and recall, though specificity is improved for patients not documented as male. No differences in metrics are statistically significant at the $p=0.05$ significance level, as measured over 1000 bootstrap samples of each subset.}
    \label{tab:stratmetrics}
\end{table*}

\section{Demographic Performance Differences}
\label{app:demographics}
We compute metrics for each demographic subgroup, as shown in \autoref{tab:stratmetrics}. No performance differences were statistically significant at the $p=0.05$ significance level, as we lack adequate sample size to demonstrate robust differences in model performance (see \autoref{tab:demographics}). Both white and non-male patients had lower AUROC than other groups, with lower sensitivity and higher specificity, implying that the $LR_{B+M+N}$ model is less able to identify patients who experienced extubation failure in these groups. While roughly one third of our dataset is non-male, white patients make up more than 60\% of our data, meaning performance differences are not related to inadequate representation in training data.

\section{Temporal Generalizability}
\label{app:temporal_gen}

We assess the intertemporal generalizability of the clinical note variables on an out-of-domain test set. First, we train a logistic regression model on encounters from 2021-2022, including base, medication, and extracted features. Second, we train a model on the same set of encounters, but omit the extracted features. We then assess the fit of both models on a subset of our test split containing encounters from 2023. In this experiment, we determine that performance (based on AUROC) is similar between the model including extracted features (0.751) and that not including such features (0.756). To ensure this performance was due to the out of domain test set, we also sample a randomly sampled subset of our test split of identical size to the 2021-2022 encounters subset, and measure performance over the 2023 test set for models fit using the extracted features (0.761) and one not using such features (0.754). The 2021-22 subset of our train set includes 1866 examples (72\% of our training set); the 2023 test subset includes 193 examples (30\% of our test set).

The distributions of outcomes among the 2021-22 patients are quite different from those among the 2023 patients: in the 2021-22 training subset, the extubation failure rate was 21\% whereas it was 13\% in 2023 (the relatively low EF rate may also explain why the models trained using smaller, out-of-domain sets attained higher AUROC than the models evaluated over all test data). Our finding demonstrates that despite the additional utility of predicting EF in-domain, rapid shifts, such as those between 2021-23 in our institution, may necessitate model retraining, and as \citet{futoma2020myth} suggest, the specific use-case of the model should guide its use to predict outcomes for new patients.

\section{Calibration Analysis}
\label{app:calibration}

\begin{figure*}[ht]
\floatconts
      {fig:calibration}
      {
      \vspace{-5mm}
      \caption{Calibration curves for the $LR_{B+M}$ and $LR_{B+M+N}$ models; both attain similar calibration, with calibration curve slopes $>1$ and Brier scores of roughly 0.14. \vspace{-3mm}
      }
      }
      {%
        \subfigure[Calibration: $LR_{B+M}$]{\label{fig:lrbm_cal}%
          \includegraphics[width=0.45\linewidth]{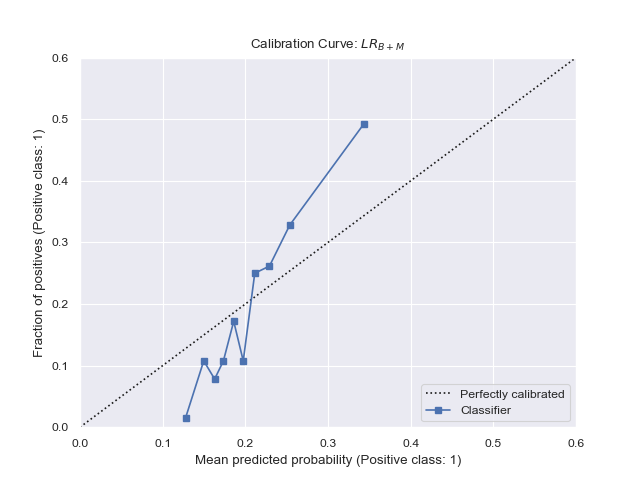}}%
        \qquad
        \subfigure[Calibration: $LR_{B+M+N}$]{\label{fig:lrbmn_cal}%
          \includegraphics[width=0.45\linewidth]{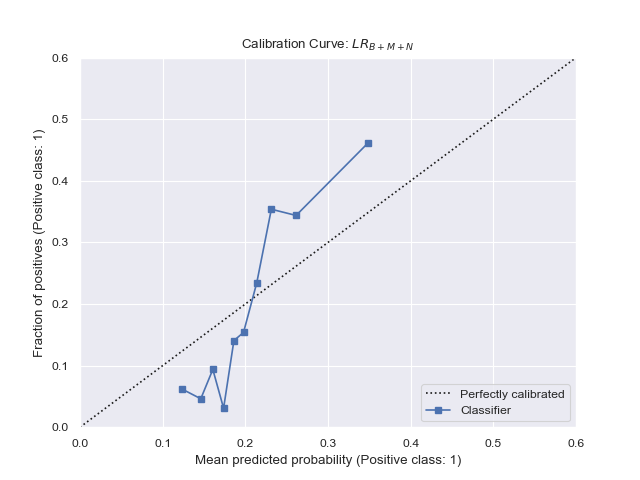}}
      }
\end{figure*}

\autoref{fig:calibration} depicts the calibration curves of the $LR_{B+M}$ and $LR_{B+M+N}$ models. Each of these curves has a slope greater than 1, indicating that each model’s predictions vary less than the underlying probability of extubation failure. Brier scores were roughly 0.14 for both variants, indicating that inclusion of features from the RT notes does not significantly affect performance.

\section{Decision Curve Analysis}
\label{app:dca}

\begin{figure}[ht!]
    \centering
    \includegraphics[width=0.9\linewidth]{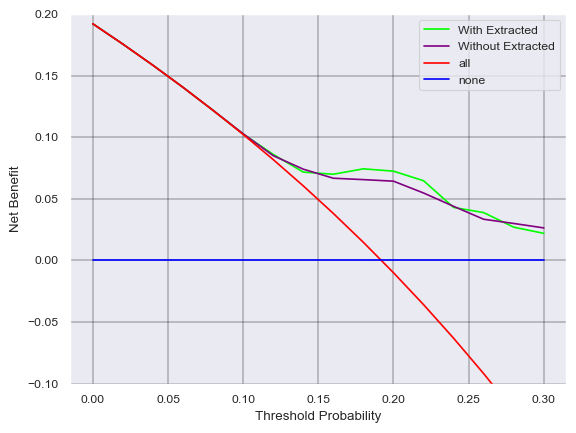}%
    \caption{Decision curve analysis comparing the net benefit of the $LR_{B+M+N}$ model to the $LR_{B+M}$ model. The $LR_{B+M+N}$ model (``With Extracted'' in the plot), which contains respiratory note features, attains a net benefit of roughly 0.01 true positives higher than the $LR_{B+M}$ model (``Without Extracted'' in the plot) at thresholds between 0.17 and 0.22.
    }%
    \label{fig:dca}%
\end{figure}

We apply Decision Curve Analysis \citep{vickers2006decision} to assess the net benefit of including additional extracted features when training EF prediction models in terms of additional true positives. This analysis indicates that including extracted features attains increased true positives per false positive at thresholds between 0.17 and 0.22, with net benefit similar at higher or lower thresholds. \autoref{fig:dca} depicts the decision curves for the $LR_{B+M+N}$ and $LR_{B+M}$ models.

\end{document}